\documentclass[letterpaper]{article} % DO NOT CHANGE THIS
\usepackage{aaai2026}  % DO NOT CHANGE THIS
\usepackage{times}  % DO NOT CHANGE THIS
\usepackage{helvet}  % DO NOT CHANGE THIS
\usepackage{courier}  % DO NOT CHANGE THIS
\usepackage[hyphens]{url}  % DO NOT CHANGE THIS
\usepackage{graphicx} % DO NOT CHANGE THIS
\usepackage{natbib}  % DO NOT CHANGE THIS AND DO NOT ADD ANY OPTIONS TO IT
\usepackage{caption} % DO NOT CHANGE THIS AND DO NOT ADD ANY OPTIONS TO IT
\usepackage{algorithm}
\usepackage{algorithmic}

\usepackage{newfloat}
\usepackage{listings}
\DeclareCaptionStyle{ruled}{labelfont=normalfont,labelsep=colon,strut=off} % DO NOT CHANGE THIS
\floatstyle{ruled}
\newfloat{listing}{tb}{lst}{}
\floatname{listing}{Listing}
\title{FreeAskWorld: An Interactive and Closed-Loop Simulator for Human-Centric Embodied AI}
\author{
    Yuhang Peng\textsuperscript{\rm}\equalcontrib,
    Yizhou Pan\textsuperscript{\rm}\equalcontrib,
    Xinning He\textsuperscript{\rm}\equalcontrib,
    Jihaoyu Yang\textsuperscript{\rm},
    Xinyu Yin\textsuperscript{\rm},
    Han Wang\textsuperscript{\rm},
    Xiaoji Zheng\textsuperscript{\rm},
    Chao Gao\textsuperscript{\rm},
    Jiangtao Gong\textsuperscript{\rm}\thanks{Corresponding author: Jiangtao Gong (gongjiangtao@air.tsinghua.edu.cn)}
}
\affiliations{
    \textsuperscript{\rm}Institute for AI Industry Research, Tsinghua University\\Email: gongjiangtao@air.tsinghua.edu.cn
}

\usepackage{bibentry}
\begin{document}

\maketitle

\begin{abstract}
As embodied intelligence emerges as a core frontier in artificial intelligence research, simulation platforms must evolve beyond low-level physical interactions to capture complex, human-centered social behaviors. We introduce FreeAskWorld, an interactive simulation framework that integrates large language models (LLMs) for high-level behavior planning and semantically grounded interaction, informed by theories of intention and social cognition. Our framework supports scalable, realistic human-agent simulations and includes a modular data generation pipeline tailored for diverse embodied tasks. To validate the framework, we extend the classic Vision-and-Language Navigation (VLN) task into a interaction enriched Direction Inquiry setting, wherein agents can actively seek and interpret navigational guidance. We present and publicly release FreeAskWorld, a large-scale benchmark dataset comprising reconstructed environments, six diverse task types, 16 core object categories, 63,429 annotated sample frames, and more than 17 hours of interaction data to support training and evaluation of embodied AI systems. We benchmark VLN models, and human participants under both open-loop and closed-loop settings. Experimental results demonstrate that models fine-tuned on FreeAskWorld outperform their original counterparts, achieving enhanced semantic understanding and interaction competency. These findings underscore the efficacy of socially grounded simulation frameworks in advancing embodied AI systems toward sophisticated high-level planning and more naturalistic human-agent interaction. Importantly, our work underscores that \textbf{interaction itself serves as an additional information modality}.
\end{abstract}

\begin{links}
    \link{Code}{https://github.com/AIR-DISCOVER/FreeAskWorld}
    \link{Datasets}{https://huggingface.co/datasets/Astronaut-PENG/FreeAskWorld}
    % \link{Extended version}{https://aaai.org/example/extended-version}
\end{links}

\section{Introduction}

Understanding and following human-generated navigation instructions is a critical capability for embodied AI agents operating in real-world environments. Vision-and-Language Navigation (VLN) tasks, which require agents to interpret natural language directives and traverse complex visual scenes, have emerged as a central research area integrating computer vision, natural language processing, and robotics. Benchmarks such as Room-to-Room (R2R) and its variants have significantly advanced progress in this domain. However, existing VLN systems are still constrained by three core limitations.

First, most methods rely on static, one-shot instructions provided at the beginning of a navigation episode, limiting the agent’s ability to handle dynamic goals or engage in multi-turn interactions. Second, current VLN frameworks often decouple high-level planning from social intention modeling, resulting in agents that cannot interpret socially salient cues or perform context-aware, human-like behaviors. Third, despite improvements in realism, simulators supporting VLN research—such as Grutopia—frequently lack complex, interactive, and dynamic elements such as moving pedestrians, social interactions, and real-world environmental variation, making them insufficient for modeling socially grounded human-agent communication.

Simultaneously, the development of generative AI and large language models has opened new possibilities for modeling high-level behaviors in simulation environments. Recent works have demonstrated that LLM-driven agents can generate diverse goals, simulate social interactions, and perform role-based tasks in virtual societies. Nevertheless, these approaches often remain disconnected from embodied navigation research, and lack mechanisms for real-time closed-loop interaction grounded in semantic and spatial understanding.

To bridge this gap, we propose FreeAskWorld—an interactive and closed-loop simulation framework designed for human-centric embodied AI. Our system leverages large language models for high-level intention modeling, semantic instruction generation, and naturalistic human behavior simulation. Grounded in theories of social behavior and intention, FreeAskWorld enables dynamic and contextually rich interactions between AI agents and human-like avatars within photorealistic 3D environments.

We further extend the classical VLN paradigm by introducing the Direction Inquiry Task, a novel benchmark that allows agents to proactively seek help and adjust their navigation based on new information, thereby evaluating higher-order capabilities such as self-assessment, social interaction, and real-time adaptation. To support this task, we construct and release the FreeAskWorld Dataset, a large-scale synthetic dataset featuring diverse human avatars, reconstructed urban environments, dynamic vehicles, and six categories of synthetic data (e.g., dialog histories, panoramic RGB, occupancy maps). The dataset includes over six hours of interactive simulation data and supports both open-loop and closed-loop evaluation.

Our work aims to fill this gap by introducing:

\begin{itemize}

    \item An interactive, LLM-driven simulation framework featuring dynamic and realistic human agents with high-level intention modeling, semantic interaction control, and a synthetic data generation pipeline.
    
    \item A novel benchmark task for direction inquiries that enables controllable and extensible devaluation of human-centric social navigation and interaction.
    
    \item We evaluated several representative VLN models, along with a human baseline, to validate the rationality of the proposed task and simulator, as well as the effectiveness of the generated data.
\end{itemize}

\section{Related Work}
\subsection{Simulating Human Behavior in Social Navigation}
% Core behavioral challenges in social navigation span from individual-level proxemics to macro-level social signals that govern multi-agent interactions \cite{rios2015proxemics,10.1145/3583741}. Prior work has modeled aspects such as personal space, intention estimation, and crowd dynamics \cite{grzeskowiak2021crowd,tsoi2022sean}, enhancing our understanding of human behavior in robot-populated environments. 
% HuNavSim advances agent-level interaction realism by modeling diverse human responses to robots—such as feeling curious or scared—via approach or avoidance\cite{perez2023hunavsim}. Arena 4.0 further expands simulation capabilities by covering a broader range of scenarios and distinguishing among human-object, human-human, and human-robot interactions, which enables more nuanced evaluation of social navigation behaviors across platforms\cite{shcherbyna2024arena}. However, while these systems excel at simulating human behavior, they typically lack mechanisms to emulate how humans generate verbal instructions, which is essential for developing embodied agents capable of engaging in more human-like, socially appropriate interactions. 

Core behavioral challenges in social navigation span from individual-level proxemics to macro-level social signaling in multi-agent contexts \cite{rios2015proxemics,10.1145/3583741}. Traditional simulation platforms have primarily relied on low-level physical behavior models to simulate crowd dynamics and agent-level reactions to robots \cite{grzeskowiak2021crowd,tsoi2022sean,vuong2024habicrowd,perez2023hunavsim}. While effective for modeling local interactions and physical constraints, these approaches often miss the dynamic and contextual nuances of societal-level human behavior.
To address this, recent work has turned to generative AI to simulate high-level, context-aware behaviors\cite{wang2024survey, xi2025rise, piao2025agentsociety}. For instance, MARPLE introduces a hierarchical structure that decomposes behavior into missions, subgoals, and atomic actions \cite{jin2024marple}. Recent simulators, such as Virtual Community\cite{zhou2025virtual}, MetaUrban\cite{wu2024metaurban}, and Grutopia\cite{wang2024grutopia}, push this further to urban-scale world by modeling open-ended social interactions among autonomous generative agents.

Building on these advances, we propose a simulation framework that integrates LLMs for high-level planning and semantically grounded interaction, informed by theories of intention and social behavior. Combined with Unity’s animation engine and asset library, our approach enables realistic, scalable, and interactive human-centric simulations.

\subsection{Human-Centric Language Use in Navigation}
% Understanding and following human-generated navigation instructions is a crucial aspect of embodied navigation systems \cite{kollar2010toward,wang2023lana}.
% Humans navigation often rely on symbolic spatial information\cite{talbot2020robot}.
Wayfinding through verbal directions is a complex and dynamic process, shaped by the direction giver, the environment, the task at hand, and the recipient\cite{hund2008role}. Two common strategies in verbal instructions are the route perspective, using egocentric cues like landmarks and left or right turns; and the survey perspective, based on map-like descriptions involving distances, street names, and cardinal directions. Navigation style from the direction giver's perspective is primarily influenced by socio-spatial background and gender \cite{galea1993sex,lawton1996strategies, lawton2001gender,kato2003individual}. For example, irregular European layouts tend to favor landmark-based navigation, while grid-based American cities rely more on street names\cite{hund2012impact}. Women usually use more spatial references, provide longer instructions, and incorporate hedging expressions\cite{sing2011gender}.

\subsection{Vision Language Navigation}
Vision-and-Language Navigation (VLN) is a foundational problem in embodied AI, where an agent interprets natural language instructions to navigate visually rich 3D environments. This interdisciplinary task integrates computer vision, natural language understanding, and robotics, driving progress toward agents capable of understanding and following human-like directions in realistic settings.

The classical VLN task was introduced by the Room-to-Room benchmark~\cite{anderson2018vision}, which provides natural language navigation instructions grounded in real-world indoor panoramic scenes. However, R2R relies on discrete navigation graphs, limiting realism and fine-grained spatial interaction. REVERIE~\cite{qi2020reverie} extends R2R by introducing object grounding, but still lacks continuous motion and physical embodiment. R2R-CE~\cite{krantz2020beyond} addresses this by enabling continuous action spaces with real-time physics, introducing challenges like low-level control and precise localization. Talk2Nav~\cite{vasudevan2021talk2nav} further advances VLN with long-range instruction following using dual attention mechanisms and spatial memory modules.

Despite progress, VLN systems face three key limitations: (1) reliance on static, one-shot instructions, reducing adaptability to dynamic environments and multi-turn interactions; (2) lack of integration between high-level planning and social intention modeling, limiting agents’ ability to interpret social cues and perform context-aware behaviors; and (3) simulators that fail to capture the complexity of real-world environments with dynamic human agents, moving vehicles, and socially grounded interactions~\cite{wang2024grutopia}.

To address these gaps, we propose a socially aware VLN framework that integrates realistic simulation environments with dynamic, human-like instruction generation. Leveraging large language models and behavior priors, our system enables socially grounded, real-time interactions, unifying language understanding, motor control, and social reasoning in a continuous embodied setting.

\begin{figure*}[htbp]
    \centering
    \includegraphics[width=1.0\textwidth]{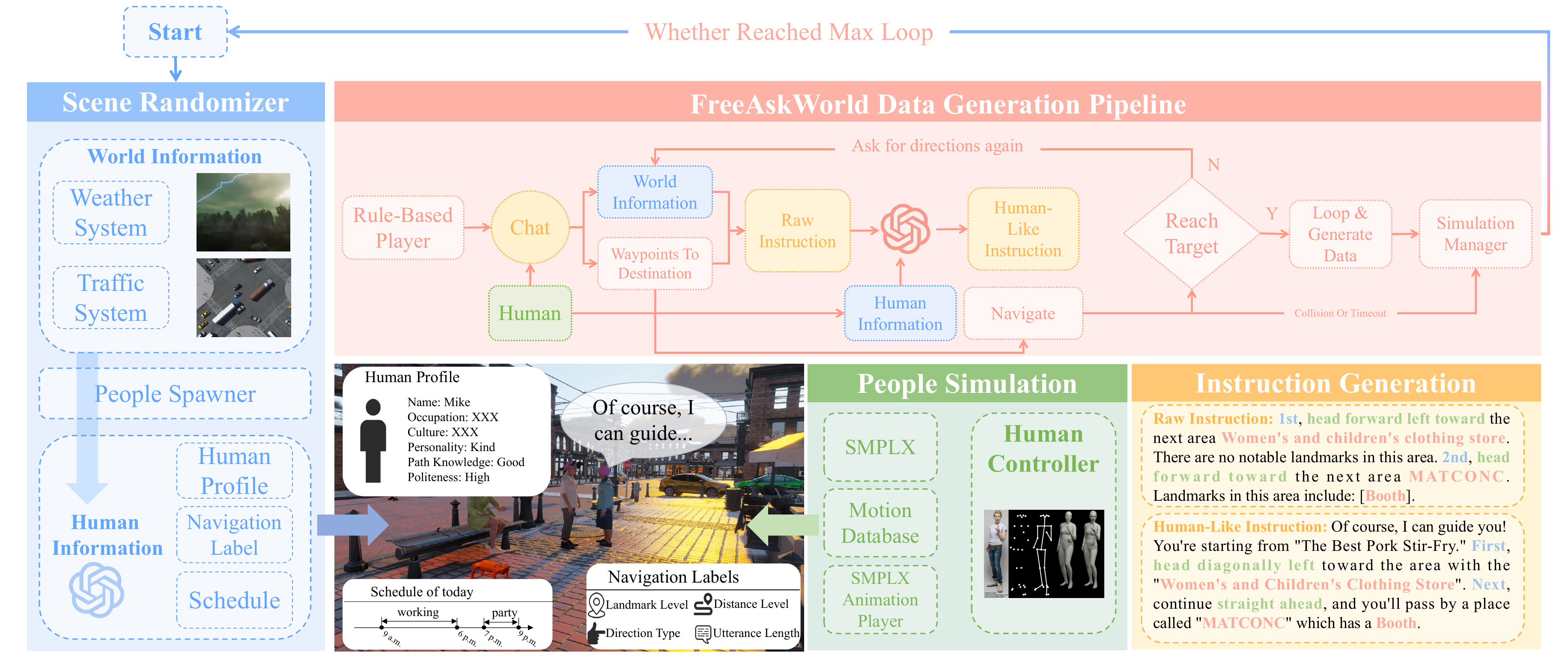}
    \caption{An overview of the FreeAskWorld framework and its data generation pipeline is presented. The system incorporates scene randomization techniques to enhance environmental diversity. The data generation module produces the FreeAskWorld dataset through this pipeline. People simulation module for modeling realistic human behaviors within virtual scenes, as well as an instruction generation module for producing navigation directives used in the Direction Inquiry Task within the simulator.}
    \label{fig:FreeAskWorldMainDesign}
\end{figure*}

\section{FreeAskWorld: Simulator Design}
\subsection{Motivation and Overview}
% 为了支持高层次人机交互任务（High-Level Human-computer Interaction Tasks）的仿真研究，需要具备更高保真度的人类行为模拟与复杂社会系统建模能力。现有大多数仿真环境往往聚焦于物理层面或任务层面的交互，缺乏对社会行为、环境因素及长期动态的真实刻画，难以支撑更复杂的交互策略研究。为此，我们提出 FreeAskWorld 仿真平台，旨在构建一个具备社会行为建模能力、开放交互接口与多系统融合的高保真虚拟社会环境。FreeAskWorld 模拟器通过引入真实社会结构与人类行为规则，实现了多层次的行为建模与决策流程。除此之外，平台中集成了包括交通系统、天气系统、日常作息机制在内的多个子系统，以增强环境的动态性和不可预测性，从而更好地模拟现实世界的复杂性。这种设计使得研究者能够在一个更接近真实社会的框架下，验证各类高层次交互策略（如自然语言协商、任务协调、社会导航、长期信任建构等）的有效性与泛化能力。此外，FreeAskWorld 支持大规模多智能体部署与行为追踪，通过采用模块化接口设计，FreeAskWorld 能够灵活集成语言模型、多模态感知模块等各类智能体模型，为研究类人社会中智能体的可持续互动与自主学习能力提供了良好的实验平台。
To facilitate the simulation of high-level human-computer interaction tasks, it is essential to achieve high-fidelity modeling of human behavior as well as complex social systems. Existing simulation environments primarily focus on physical-level or task-level interactions, often lacking realistic representations of social behavior, environmental dynamics, and long-term temporal evolution. This limitation hinders the development and evaluation of advanced interaction strategies.

To address this gap, we propose FreeAskWorld, a high-fidelity simulation platform designed to model human-like societies with structured social behavior, open interaction interfaces, and integrated multisystem dynamics. FreeAskWorld incorporates real-world social structures and behavioral norms to enable multilevel behavior modeling and decision-making processes. In addition, the platform integrates several subsystems, such as transportation, weather, and daily activity cycles, to introduce environmental variability and unpredictability, thus enhancing the realism of the simulated world.

This design lays the foundation for evaluating a wide range of high-level interaction strategies in environments that closely approximate real-world social contexts. In the current stage, we focus on direction inquiries as a representative interaction task. However, FreeAskWorld is designed to support, in future developments, more complex interactions such as natural language negotiation, task coordination, social navigation, and long-term trust building.

Furthermore, FreeAskWorld supports large-scale multi-agent deployment and behavior tracking. Through a modular interface architecture, it enables seamless integration of language models, multimodal perception modules, and other intelligent components, offering a flexible and extensible platform for studying sustained interaction and autonomous learning in human-like societies.

\subsection{People Simulation}
\begin{figure*}
    \centering
    \includegraphics[width=0.9\textwidth]{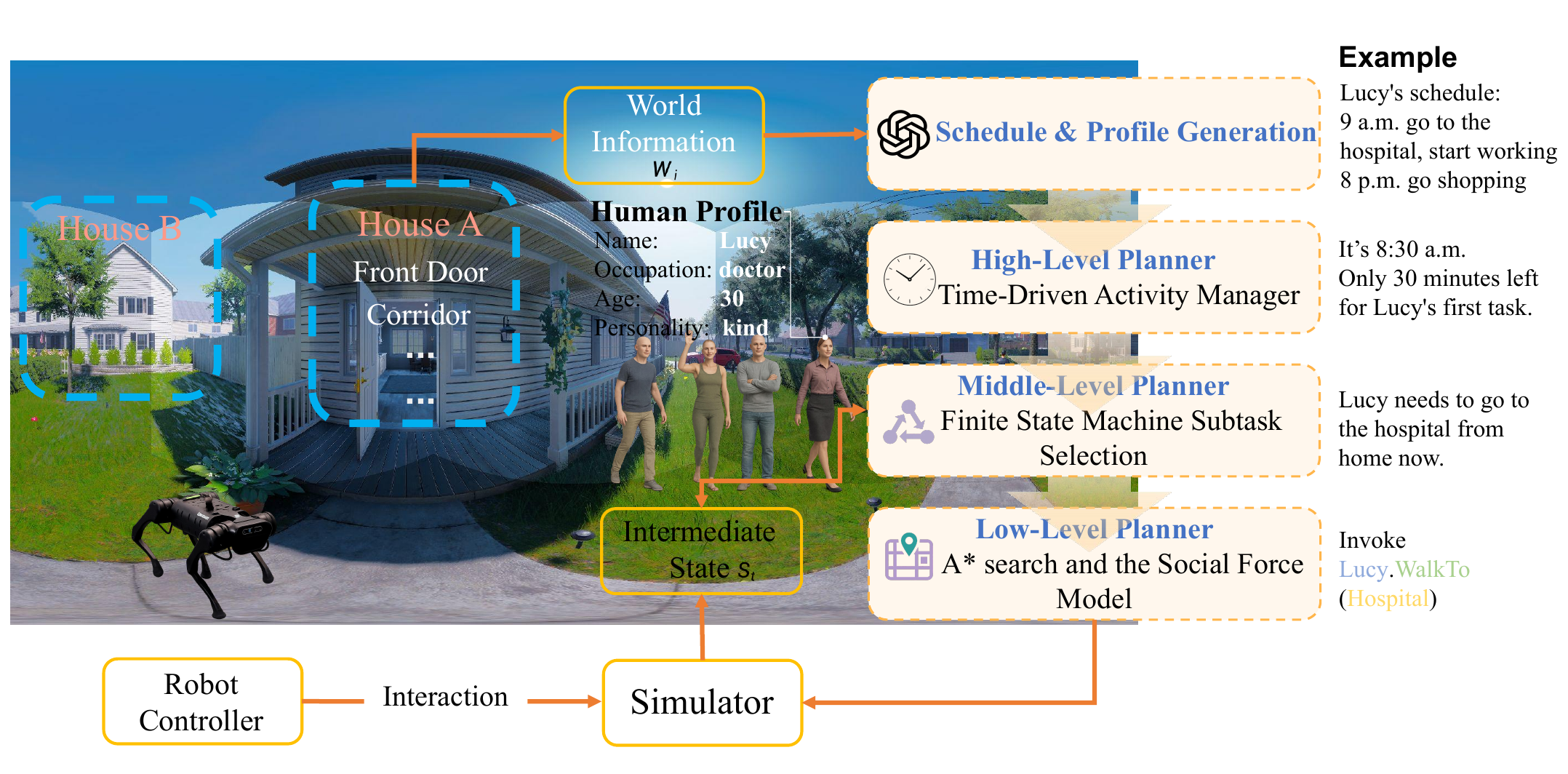}
    \caption{The People Simulation works as follows: a large language model creates diverse character profiles and schedules, which the high-level planner uses to select activities based on time. These activities are passed to the middle-level planner, which breaks them into subtasks and manages them with a finite state machine. Low-level planners handle navigation and basic behaviors, using A* for global paths and the Social Force Model to avoid obstacles locally.}
    \label{fig:PeopleSimulationDesign}
\end{figure*}

This section introduces the People Simulation module, which encompasses avatar modeling, profile and schedule generation, navigation style synthesis, animation control, and appearance variation.  
An overview of the module is illustrated in Figure~\ref{fig:PeopleSimulationDesign}.
\subsubsection{Avatar Models}
The visual and kinematic realism of human agents significantly impacts the credibility of Sim2Real navigation systems. Many navigation simulators still rely on static 3D character models with minimal animation diversity\cite{li2021igibson,vuong2024habicrowd}. HA-VLN takes a step forward by combining SMPL-based body models with AI-based action planning, enhancing the naturalness of character motion \cite{dong2025ha}.

% 我们使用SMPLX作为Avatar的建模，处于对性能的考虑，使用blender的Decimate Modifier对模型进行降面，模型的材质纹理使用MLLM生成不同职业性别的外观。
\subsubsection{Profile And Schedule Generation}
We employ a two-stage generation framework to create diverse and realistic human agents. First, character profiles are synthesized incorporating demographic and contextual attributes, such as age, culture, and occupation. In the second stage, given the character profile and a static scene layout, we generate a corresponding daily schedule, which includes multiple activities with temporal segmentation and location assignments sampled from available destinations within the environment. 
% All schedule entries are generated by prompting GPT-4 to produce semantically coherent routines aligned with the agent’s role, personality, and context.

\subsubsection{Navigation Style Generation}
We incorporate regional familiarity and personality into the role profile and classify navigation style by four key features: landmark use, direction type, distance description, and utterance length. The findings of the literature are encoded as a knowledge base, enabling the LLM to generate contextually grounded navigation labels and instructions based on the role profile, task, and geological situation.

\subsubsection{Animation Database}
To support consistent embodiment and context-sensitive behaviors, we adopt MotionX\cite{lin2023motionx} as the SMPL-X animation library and structure its motions into high-level categories and subcategories, allowing the selection of semantically relevant motions for each activity.
% 并编写了SMPLX模型动画驱动插件，以此可以随意使用动画库里面的所有动画，同时我们也为动画控制器加入了Blending功能，以实现动作之间的平滑过渡，提高动画的保真度和人物的真实度。
We also developed a custom SMPL-X animation driver plugin that allows seamless access to the full range of animations within the library. To enhance animation fidelity and character realism, we integrated Blending functionality into the motion controller, enabling smooth transitions between actions.

\subsubsection{Apperance Variation}
\textit{Method 1:}  We propose a new framework that uses Multimodal Large Language Models (MLLMs) to create diverse virtual human appearances controlled by natural language. By combining UV mapping with semantic profiles (such as gender, occupation, and ethnicity), we build a comprehensive pipeline for texture generation guided by language, enabling scalable and consistent material creation across a variety of scenarios. At the mesh level, we introduce variation by randomly modifying the shape parameters of the SMPL-X model, producing diverse body types characterized by differences in height, weight, and proportions. This approach, coupled with semantic control over beta shapes properties, enhances both the realism and expressiveness of human simulations.
\textit{Method 2:} We leverage the Synbody~\cite{yang2023synbody} dataset to generate SMPL-X models, which incorporate clothing, footwear, hair, and other detailed features, offering a more immersive and realistic virtual experience. This approach enhances the fidelity of modeling human appearance.

Detailed examples are provided in the Technical Appendix.

\begin{figure}
    \centering
    \includegraphics[width=0.40\textwidth]{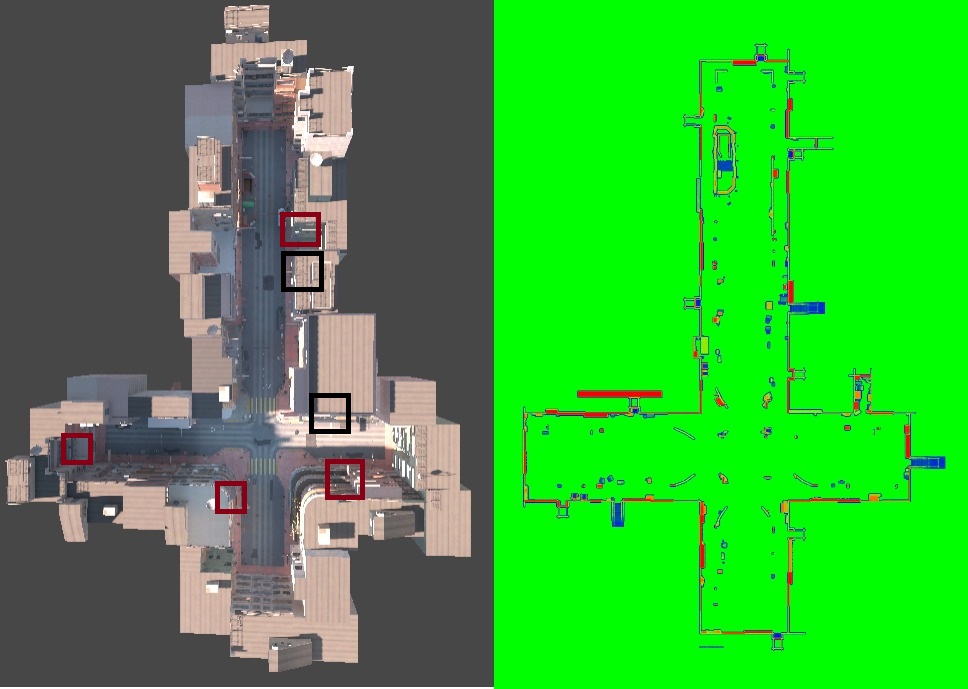}
    \caption{Comparison Between the Original Mesh Model and the Generated Occupancy Heatmap. The black and red bounding signs represent the same store A and B in different positions, a layout designed to assess the navigation capabilities of humans or robots in complex environments.}
    \label{fig:OccupancyMap} % 可用于交叉引用
\end{figure}

\subsection{Other System Functions}

\subsubsection{Occupancy Map Generation}
We generate a 2D occupancy heat map based on voxels to support navigation, mapping, and interaction. The environment is divided into 3D voxels within a set region. For each voxel, occupancy is estimated probabilistically through random sampling and collision checks, producing a soft occupancy value.

To improve reliability, we repeat the sampling multiple times and average the results. The occupancy map is then refined using filtering, morphological operations, and noise removal techniques. Finally, the 3D map is projected onto a 2D plane to create top-down occupancy maps, which can be exported as heatmaps or binary grids. Full implementation details are in the Technical Appendix.

An example is shown in Fig.~\ref{fig:OccupancyMap}.

\subsubsection{Weather System}
% 我们使用Unity的Enviro 3 - Sky and Weather插件实现昼夜以及天气的模拟，通过对天气系统的模拟可以提升数据采集的泛化性和仿真器其他部分的真实性。
We use a dynamic sky and weather system to simulate day-night cycles and various weather conditions like rain and fog. This improves simulation realism and data diversity, helping models trained on this data generalize better under different lighting and visibility.

\subsubsection{Traffic Simulation}
% 我们使用Unity的Urban Traffic System插件，这个插件节点化建模行驶路线和驾驶规则，实现对人类社会中的车辆和交通系统行为的模拟，用于增加仿真器的拟真度和环境动态物体的复杂度。
We integrate a traffic simulation system that models vehicle movement and traffic rules using route graphs. This adds realistic vehicle behaviors and dynamic complexity to the environment, supporting more challenging perception and planning tasks.

\subsubsection{Robot Simulation}
% 机器人的导航方面使用A*算法做全局规划，使用SFM社交力模型实现局部避障，同时我们实验了数据采集时是否应用SFM模型时的成功率，见表格\ref{tab:RobotSimulation}。同时，我们仿真器支持机器人动力学的模拟，使用articulation组件来实现各种复杂机器人关节的模拟，并且我们提供了一个四轮机器人可以供后续机器人底层运动控制的模拟
Our system uses the A* algorithm for global path planning and the Social Force Model (SFM) for local obstacle avoidance, enabling socially-aware navigation in dynamic environments. The application of SFM during data collection significantly improves navigation success rates by effectively avoiding pedestrians and vehicles, though it results in longer trajectories due to more cautious path planning.
Furthermore, the simulator supports realistic robot dynamics through Unity's articulation components and allows seamless integration of physical robots via the URDF-Importer.

% \begin{table}[htbp]
% \centering
% \caption{Comparison of navigation performance with and without the Social Force Model. SR denotes Success Rate (\%), and TL represents the average Trajectory Length (meters).}
% \label{tab:RuleBasedNavigationExperiment}
% \renewcommand{\arraystretch}{1.2}
% \setlength{\tabcolsep}{3pt} % 默认是6pt，减小列间距
% \begin{tabular}{|c|c|c|}
% \hline
% \textbf{Method}  & \textbf{TL} & \textbf{SR $\uparrow$} \\
% \hline
% A* Only & 45.64 & 82.05 \\
% A* + SFM & 55.51  & \textbf{90.37}\\
% \hline
% \end{tabular}
% \end{table}

\subsubsection{Synchronous Closed-loop Framework}
% 鉴于服务器环境通常无法运行非Headless模式的仿真器，我们设计并实现了一种基于WebSocket的同步闭环仿真架构。该架构既支持通过内网穿透技术实现服务器端模型与仿真器端的网络连接，完成闭环仿真，也支持在同一设备上通过端口通信方式进行数据交互。系统提供多样化的消息接口，以满足不同类型传感器数据及控制命令的传输需求。
Considering that non-headless simulators typically cannot be run in server environments, we have designed and implemented a WebSocket-based synchronous closed-loop simulation architecture. This architecture supports closed-loop simulation through network connections between the server-side model and the simulator via NAT traversal techniques, as well as data exchange via port communication on the same device. The system provides diverse message interfaces to accommodate the transmission of various types of sensor data and control commands.

\section{Direction Inquiry Task}
% 我们对VLN任务进行了改进，增加了问路环节，以评测模型的高层次任务（判断自身状态自主向外界获取信息并利用信息进行规划）的规划能力和逻辑判断能力，并且在仿真其中有大量的动态人类和车辆，以评测模型的低层次运动规划能力。
We extend the traditional Vision-and-Language Navigation (VLN) task by introducing an inquiry phase, allowing the agent to actively seek external information. This modification enables the evaluation of the model's high-level capabilities, such as self-assessment, information-seeking behavior, and planning based on acquired knowledge. Additionally, our simulation environment includes a large number of dynamic humans and vehicles, facilitating the assessment of the model’s low-level motion planning and control capabilities under realistic and dynamic conditions.

\subsection{Evaluation Metrics}
% Success Rate, SR
% Trajectory Length, TL
% Success weighted by Path Length, SPL
% Navigation Error, NE
% Oracle Navigation Error, ONE
% Oracle Success Rate, OSR
% Number of Direction Inquiries, NDI

To comprehensively evaluate the agent's performance in the Direction Inquiry Task, we employ the following standard metrics commonly used in navigation and vision-language tasks:

\begin{itemize}
\item \textbf{Trajectory Length (TL)}: The average total distance traveled by the agent, computed as  
$ \mathrm{TL} = \frac{1}{N} \sum_{i=1}^N \sum_{t=1}^{T_i - 1} d(p_i^{t+1}, p_i^t) $,  
where $T_i$ is the number of time steps in episode $i$.

\item \textbf{Success weighted by Path Length (SPL)}: Measures both success and efficiency~\cite{anderson2018vision}, defined as  
$ \mathrm{SPL} = \frac{1}{N} \sum_{i=1}^N S_i \frac{l_i^*}{\max(l_i, l_i^*)} $,  
where $S_i \in \{0,1\}$ indicates success in episode $i$, $l_i^*$ is the shortest path length to the goal, and $l_i$ is the actual path length taken by the agent.

\item \textbf{Success Rate (SR)}: The proportion of successful episodes,  
$ \mathrm{SR} = \frac{1}{N} \sum_{i=1}^N S_i $,  
where success is defined as reaching within a threshold distance $\delta$ of the goal.

\item \textbf{Navigation Error (NE)}: The average final distance between the agent and the goal,  
$ \mathrm{NE} = \frac{1}{N} \sum_{i=1}^N d(p_i^{T_i}, g_i) $.

\item \textbf{Oracle Navigation Error (ONE)}: The average minimum distance to the goal along the trajectory,  
$ \mathrm{ONE} = \frac{1}{N} \sum_{i=1}^N \min_{t=1,\dots,T_i} d(p_i^t, g_i) $.

\item \textbf{Oracle Success Rate (OSR)}: The fraction of episodes in which the agent is within $\delta$ of the goal at any point during its trajectory,  
$ \mathrm{OSR} = \frac{1}{N} \sum_{i=1}^N \mathbf{1}\!\left(\min_{t=1,\dots,T_i} d(p_i^t, g_i) \leq \delta \right) $,  
where $\delta$ is the radius of success (typically 1–3 m depending on the environment).

\item \textbf{Number of Direction Inquiries (NDI)}: The average number of inquiry actions issued by the agent per episode,  
$ \mathrm{NDI} = \frac{1}{N} \sum_{i=1}^N n_i^{\mathrm{inquiry}} $.
\end{itemize}

\begin{table}[t]
\centering
\caption{Comparison of Vision-and-Language Navigation Datasets}
\label{tab:datasets}
\scriptsize
\setlength{\tabcolsep}{9pt}
\begin{tabular}{lcccc}
\hline
\textbf{Dataset} & \textbf{Scene} & \textbf{Type} & \textbf{Inst.} & \textbf{Traj.} \\
 & & & \textbf{(words)} & \textbf{(m)} \\
\hline
Talk2Nav & Outdoor & Discrete & $\sim$70 & $\sim$60 \\
R2R & Indoor & Discrete & $\sim$29 & $\sim$20 \\
REVERIE & Indoor & Discrete & $\sim$18 & $\sim$12 \\
ScaleVLN & Indoor & Discrete & $\sim$29 & $\sim$10 \\
NavRAG & Indoor & Discrete & $\sim$25 & $\sim$8 \\
GSA-R2R & Indoor & Discrete & $\sim$35 & $\sim$12 \\
HA-R2R & Indoor & Discrete & $\sim$29 & $\sim$10 \\
NaVid & Indoor & Continuous & $\sim$20 & $\sim$25 \\
DynamicVLN & Outdoor & Continuous & $\sim$28 & $\sim$400 \\
VLN-Video & Outdoor & Continuous & $\sim$90 & $\sim$200 \\
FreeAskWorld & In/Outdoor & Continuous & $\sim$148 & $\sim$56 \\
\hline
\end{tabular}
\end{table}

\section{FreeAskWorld Dataset}
\subsection{Dataset Overview}
We compare our dataset with other existing VLN datasets, including Talk2Nav~\cite{vasudevan2021talk2nav}, R2R~\cite{anderson2018vision}, REVERIE~\cite{qi2020reverie}, ScaleVLN~\cite{wang2023scaling}, NavRAG~\cite{wang2025navrag}, GSA-R2R~\cite{li2024human}, HA-R2R~\cite{li2024human}, NaVid~\cite{zhang2024navid}, DynamicVLN~\cite{sun2025dynamicvln}, and VLN-Video~\cite{li2024vln}. Table~\ref{tab:datasets} summarizes their characteristics.

\subsection{Dataset Validation}

To validate the data in our dataset, we conduct a human baseline evaluation in Figure.~\ref{tab:ClosedLoopResults}. In this process, Experimenters verify the alignment between the LLM-generated instructions and the intended destination. The generated text should exhibit qualities of being coherent, human-like, and easy to understand. This approach ensures that the instructions are both reasonable and effective in guiding the navigation process.

\subsection{Data Generation Pipeline}
% 首先进行仿真器环境的初始化与环境随机化（天气时间等），开始后数据采集器会主动寻找人并进行交谈，交谈后会得到LLM生成的仿照真人会给的指路数据，然后数据采集器会以社交导航的方式导航到指定地点，中途超时未到达地点还会寻找附件的人进行下一次问路，到达地点后标志数据采集成功，然后记录下来文本以及图像和合成数据。
The data generation pipeline begins with the initialization of the simulation environment, including randomization of environmental conditions such as weather and time of day to promote diversity and robustness. Once initialized, the data collection agent actively searches for nearby human agents and initiates an interaction to request navigational assistance. The responses are generated by a large language model (LLM) to simulate realistic, human-like instructions.

Following the dialogue, the agent navigates to the specified destination using a socially compliant navigation strategy that accounts for both static and dynamic obstacles. If the agent fails to reach the destination within a predefined time threshold, it will initiate another round of inquiry with nearby human agents to update its goal. Upon successful arrival at the target location, the episode is marked as successful, and all relevant data—including dialogue transcripts, panoramic and perspective RGB images, and associated synthetic annotations—are recorded for training and evaluation purposes

% 本数据集累计采集6小时数据，大小为多少GB。
\subsection{Sensor Configuration}
% 我们设置了6个视场角为90°的摄像头，以1 Hz的频率从同一位置进行全景图像采集，每个摄像头仅在朝向上存在旋转差异。采集过程中同时记录了采集器在世界坐标系下的位置和旋转信息，用于与合成图像数据与世界信息进行位置变换关联。
We configured six cameras, each with a 90 ° field of view (FOV), positioned at the same spatial location but oriented in different directions to collectively capture panoramic images at a frequency of 1 Hz. During the data acquisition process, we simultaneously recorded the position and orientation of the camera rig in the world coordinate system. This information enables the transformation and alignment between the synthesized image data and the global scene context.

\subsection{Trajectory Generation}
% 我们在仿真场景中设置了随机起点和目标点，并执行问路行为以获取导航指引。随后，智能体采用规则化路径规划算法进行导航，在此过程中会避让静态和动态障碍物，实现具有社交约束的路径生成。所产生的轨迹可作为预训练阶段供模仿学习使用的专家示范数据。
In the simulated environment, we randomly assign start and target positions, followed by an inquiry-based behavior to obtain navigational guidance. The agent then navigates to the goal using a rule-based path planning algorithm, which accounts for static and dynamic obstacle avoidance in a socially compliant manner. The resulting trajectories serve as expert demonstrations for imitation learning.

\begin{figure}
    \centering
    \includegraphics[width=0.40\textwidth]{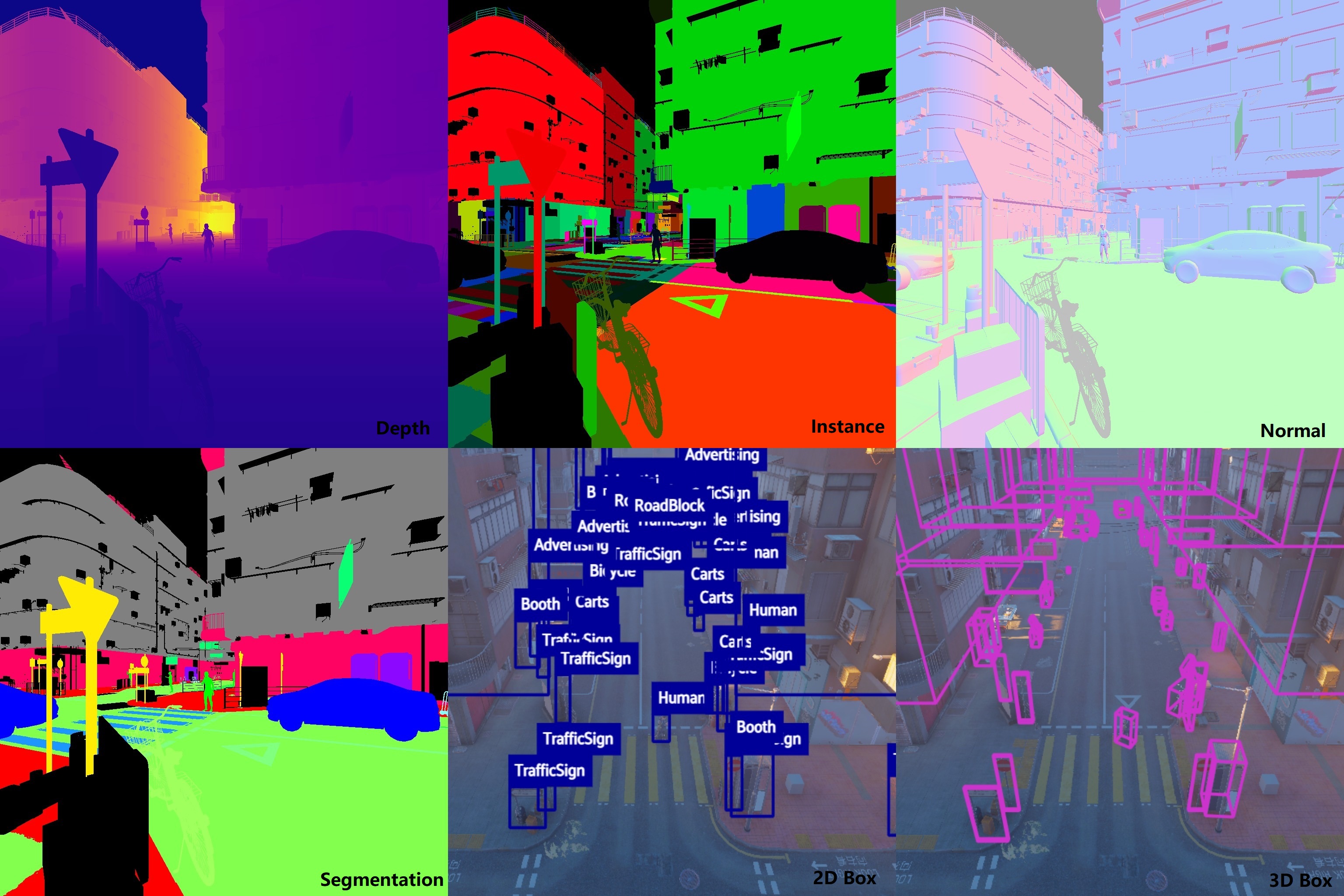}
    \caption{Six Main Types of Synthetic Data}
    \label{fig:SyntheticData} % 可用于交叉引用
\end{figure}

\subsection{Data Composition}
% 我们使用Unity的Perception包采集合成数据，六类合成任务数据及其他数据和16类关键物体类型，数据集还有场景重建
% bbox 3D
% bbox 2D
% instance
% semantic
% surface normal
% depth
% panorama and panorama's six perspective rgb img
% HCI Data(instruction, etc)
% 2D occupancy heatmap
% world informatin
We used Unity Perception\cite{borkman2021unity} to construct a rich and diverse synthetic dataset that encompasses multiple types of annotations and data modalities. The dataset is designed to support a wide range of vision and navigation tasks and includes both dense per-frame annotations and global scene-level metadata. Specifically, the dataset comprises:

\begin{itemize}
    \item \textbf{Visual annotations}: 2D/3D bounding boxes, instance and semantic segmentation.
    \item \textbf{Geometric annotations}: depth maps and surface normal maps for scene geometry.
    \item \textbf{Visual observations}: panoramic RGB images and six 90$^\circ$ perspective views.
    \item \textbf{Interaction data}: natural language instructions, dialog histories, and agent trajectories.
    \item \textbf{Spatial representations}: 2D occupancy heatmaps for mapping and localization.
    \item \textbf{Environment metadata}: map boundaries, semantic regions, and other contextual information.
\end{itemize}
As shown in Figure~\ref{fig:SyntheticData}, the six main types of synthetic data are summarized.

Additionally, the dataset includes annotations for 16 key object categories commonly encountered in human-centered environments, such as vehicles, pedestrians, and street furniture. 

\subsection{Scene Reconstruction}
% 根据2D occupancy heatmap作为静态物体的重建，3D box记录动态物体的位置信息，再根据world information可以重建整个仿真器的场景环境，并可以根据这个环境进行开环测试，进行后续更多样的导航或者人机交互任务。
Based on the 2D occupancy heatmaps that encode the layout of static elements, along with the 3D bounding boxes that capture the positions of dynamic entities, the simulation environment can be accurately reconstructed. By further integrating the provided world information, it becomes possible to generate a comprehensive digital twin of the scene. This reconstructed environment enables open-loop evaluations similar to those in the nuScenes dataset~\cite{caesar2020nuscenes}, and is particularly well suited for unstructured settings as in FreeAD~\cite{peng2025bench2freead}, and supports a broad spectrum of downstream tasks including navigation planning, behavior prediction, and human-computer interaction studies.

\section{Experiments}
%\subsection{Datasets}
% 我们在FreeAskWorld上进行训练微调模型，并在FreeAskWorld的开环测试集上跑开环实验，在FreeAskWordl的闭环测试集与仿真器上跑闭环实验
We conduct model training and fine-tuning on the FreeAskWorld dataset. For evaluation, open-loop experiments are performed in the OpenAskWorld open-loop test split, while closed-loop experiments are conducted in both the closed-loop test split and within the simulator environment.These experiments serve to validate the effectiveness of our dataset and simulator.

\subsection{Baselines}
% Human
% ETP-Nav
% BEV-BERT
% 我们在 FreeAskWorld 数据集上进行了开环与闭环两种实验评估方式：
% 在开环评估中，我们使用 FreeAskWorld 数据集的开环测试集，采用 L2 距离指标，评估模型每一帧预测的轨迹与专家轨迹之间的误差。
% 在闭环评估中，我们在 FreeAskWorld 数据集的闭环测试集及其配套仿真环境中进行实验，评估指标采用 Direction Inquiry Task 中定义的任务完成率等行为级指标。
% 在基线方面，我们设置了以下三种方法用于对比：
% Human Baseline（人类表现）：作为导航任务的性能上限基准；
% ETP-Nav：一个预训练的基于指令的导航模型；
% BEV-BERT：一个基于鸟瞰图表示的空间推理模型。
% 对于 ETP-Nav 和 BEV-BERT，我们同时报告其原始预训练模型的性能以及在 FreeAskWorld 数据集上微调后的版本结果，以全面评估现有方法在新环境中的迁移能力与适应性能。

We conduct both open-loop and closed-loop evaluations on the FreeAskWorld dataset to comprehensively assess model performance.

\begin{itemize}
    \item In the open-loop setting. We adopted the L2 distance metric to measure the per-frame deviation between predicted trajectories and expert demonstrations.
    
    \item In the closed-loop setting. We employ metrics defined in the Direction Inquiry Task.
\end{itemize}

We compare several baselines as follows:

\begin{itemize}
    \item Human: Used as an upper bound reference for navigation performance.
    \item ETPNav~\cite{an2024etpnav}: A hierarchical VLN‑CE framework that performs online topological mapping, cross-modal planning with transformers, and low-level control using a rotate‑then‑forward schema augmented by obstacle‑avoidance heuristics.
    \item BEVBert~\cite{an2022bevbert}: A map‑based multimodal pre-training model that leverages hybrid topo‑metric representations to improve spatial reasoning and language-guided navigation robustness.
\end{itemize}
We evaluate ETPNav and BEVBert using both their original pretrained models and fine-tuned versions on the FreeAskWorld dataset, referred to as ETPNav-FT and BEVBert-FT.

% \subsection{Training Details}
% Models were fine-tuned from pretrained weights on a cluster with 8 NVIDIA A800 GPUs. Model selection was based on validation performance.

\subsection{Experimental Setup}
% For open-loop evaluation, we test all models on the FreeAskWorld open-loop test set.

% For closed-loop evaluation, each episode is initialized with scene conditions (weather, time of day), the agent's pose, and the first navigation instruction from the test set. This initial instruction is not counted toward the Number of Direction Inquiries (NDI), so each episode starts with NDI = 0. Given the complexity and variability of dynamic environments, we run multiple trials per episode and report averaged results for robustness.

% For the human baseline, four participants are involved: two follow only the initial instruction, and two are allowed to ask follow-up questions (each counted as one NDI). Participants control the agent using keyboard inputs to walk, interact, and complete tasks.

% For model-based methods (ETPNav, ETPNav-FT, BEVBert, BEVBert-FT), we adopt a synchronous closed-loop framework with a 1 Hz update rate (one prediction per simulated second). Models run on an RTX 3080 GPU, and the simulator runs separately on an RTX 3060. Episodes are terminated upon pedestrian/vehicle collision or exceeding 100 steps.

For open-loop evaluation, models are tested on the FreeAskWorld open-loop test set.
In closed-loop evaluation, each episode starts with scene conditions, agent pose, and the first navigation instruction, which does not count toward the Number of Direction Inquiries (NDI), starting at NDI = 0. Multiple trials per episode ensure robustness, with averaged results reported.
The human baseline involves four participants: two follow the initial instruction only, while the other two may ask follow-up questions (each counted as one NDI). Participants control the agent with keyboard inputs.
For model-based methods (ETPNav, ETPNav-FT, BEVBert, BEVBert-FT), we use a synchronous closed-loop framework with a 1 Hz update rate. Models run on an RTX 3080, and the simulator on an RTX 3060. Episodes terminate after pedestrian/vehicle collisions or after 100 steps.

\subsection{Results}
The open-loop results show that the fine-tuned models, \texttt{ETPNav-FT} and \texttt{BEVBert-FT}, achieve a $\sim$50\% reduction in L2 error compared to their base versions, with \texttt{BEVBert-FT} delivering the best overall performance.

In closed-loop experiments, the human baseline demonstrates that agents capable of querying for additional navigation instructions significantly improve pathfinding accuracy (from 40.2\% to 82.6\%
). This improvement is attributed to scene complexity, such as identical stores appearing in different directions within 50 meters (shown in Fig.~\ref{fig:SyntheticData}), where humans may occasionally experience shallow memory of directions and become disoriented. Fine-tuned models, \texttt{ETPNav-FT} and \texttt{BEVBert-FT}, show substantial improvements over their base counterparts in both Navigation Error and Oracle Navigation Error. The increase in Trajectory Length suggests enhanced familiarity with the scene and broader exploration. Although \texttt{ETPNav-FT} achieves partial destination success, overcoming the zero Oracle Success Rate observed in baseline models, its overall Success Rate remains zero. A similar trend is observed in the Social Mobile Manipulation task of InfiniteWorld~\cite{ren2024infiniteworld}, where tasks involving social interactions substantially reduce robot performance. This challenge arises from weak dynamic social navigation, pedestrian/vehicle collisions, and limitations in long-range planning, abstract reasoning, memory retention, and higher-level decision-making—areas that warrant further investigation. Additionally, \texttt{BEVBert} consistently outperforms \texttt{ETPNav} across both evaluation metrics, reinforcing its state-of-the-art performance in Vision-and-Language Navigation (VLN) and highlighting commonalities with our task.

% The human baseline results confirm that humans can effectively complete navigation tasks, achieving approximately 40\% SR even when limited to initial navigation instruction. In contrast, state-of-the-art VLN pre-trained models struggle considerably, frequently failing due to weak dynamic social navigation skills and collisions with pedestrians and cars. In addition, these models exhibit deficiencies in long-range path planning, abstract concept understanding, long-term memory retention, and higher-level reasoning abilities such as monitoring their own state and proactively requesting navigation instructions. All of these represent promising directions for future research. Additionally, \texttt{BEVBert} consistently outperforms \texttt{ETPNav} across both open-loop and closed-loop metrics, further reinforcing its state-of-the-art performance in Vision-and-Language Navigation (VLN) tasks. This also highlights the alignment between our task and certain aspects of VLN, providing additional insights into their shared characteristics.

Overall, these evaluations not only validate the effectiveness of the proposed dataset but also shed light on the strengths and limitations of current models in socially situated navigation as well as in the Direction Inquiry Task or Interaction-oriented tasks.

% \begin{table}[htbp]
% \centering
% \caption{Open-loop Evaluation Results on the FreeAskWorld Dataset}
% \label{tab:OpenLoopResults}
% \renewcommand{\arraystretch}{1.2}
% \setlength{\tabcolsep}{24pt}
% \small
% \begin{tabular}{l c}
% \hline
% \textbf{Method} & \textbf{L2~(m)$\downarrow$} \\
% \hline
% ETPNav & 0.950 \\
% BEVBert & 0.914 \\
% ETPNav-FT & 0.560 \\
% BEVBert-FT & \textbf{0.462} \\
% \hline
% \end{tabular}
% \end{table}

\begin{table}[htbp]
\centering
\caption{Closed-Loop Navigation Performance of Various Methods in the FreeAskWorld Simulator}
\label{tab:ClosedLoopResults}
\renewcommand{\arraystretch}{1.2}
\setlength{\tabcolsep}{2pt}
\small 
\begin{tabular}{lccccccc}
\hline
\textbf{Method} & \textbf{TL} & \textbf{SR$\uparrow$} & \textbf{SPL$\uparrow$} & \textbf{NE$\downarrow$} & \textbf{OSR$\uparrow$} & \textbf{ONE$\downarrow$} & \textbf{NDI} \\
\hline
Human(no ask) & 47.5 & 40.2 & 38.2 & 18.3 & 41.3 & 11.3 & 0.0 \\
Human(ask) & 59.9 & \textbf{82.6} & \textbf{71.2} & \textbf{3.49} & \textbf{82.6} & \textbf{1.63} & 0.78 \\
ETPNav & 31.2 & 0.0 & 0.0 & 32.9 & 0.0 & 28.7 & 0.0 \\
BEVBert & 14.6 & 0.0 & 0.0 & 31.0 & 0.0 & 29.0 & 0.0 \\
ETPNav-FT & 33.6 & 0.0 & 0.0 & 31.6 & \textbf{1.1} & \textbf{27.1} & 0.0 \\
BEVBert-FT & 18.7 & 0.0 & 0.0 & \textbf{30.0} & 0.0 & 28.5 & 0.0 \\
\hline
\end{tabular}
\end{table}

\section{Conclusion}
We introduce the Direction Inquiry Task to extend traditional VLN settings, emphasizing self-assessment, social interaction, and real-time adaptation. To support this, we release the FreeAskWorld Dataset, featuring diverse avatars, dynamic scenes, and rich multimodal annotations. Experiments show that VLN models fine-tuned on our dataset improve in both open-loop and closed-loop settings. However, comparisons with human performance reveal that current models still struggle with high-level reasoning and socially grounded navigation.

Importantly, our work underscores that \textbf{interaction itself serves as an additional information modality}. Intentional and structured interaction is \textbf{not only a social signal but also a crucial pathway for understanding and interpreting the physical world}, enabling agents to acquire information that static perception alone cannot provide. This highlights the broader value of socially grounded simulation in bridging the gap between embodied AI and real-world human interaction.

Future directions include tackling complex tasks like negotiation and coordination, integrating multimodal memory and perception for adaptive behaviors, and leveraging generative models for higher visual fidelity. We also plan to develop an end-to-end software solution on Steam for easier access and expand the benchmark suite to include more comprehensive metrics to evaluate embodied AI interactions.

\section{Acknowledgments}
% AAAI is especially grateful to Peter Patel Schneider for his work in implementing the original aaai.sty file, liberally using the ideas of other style hackers, including Barbara Beeton. We also acknowledge with thanks the work of George Ferguson for his guide to using the style and BibTeX files --- which has been incorporated into this document --- and Hans Guesgen, who provided several timely modifications, as well as the many others who have, from time to time, sent in suggestions on improvements to the AAAI style. We are especially grateful to Francisco Cruz, Marc Pujol-Gonzalez, and Mico Loretan for the improvements to the Bib\TeX{} and \LaTeX{} files made in 2020.

% The preparation of the \LaTeX{} and Bib\TeX{} files that implement these instructions was supported by Schlumberger Palo Alto Research, AT\&T Bell Laboratories, Morgan Kaufmann Publishers, The Live Oak Press, LLC, and AAAI Press. Bibliography style changes were added by Sunil Issar. \verb+\+pubnote was added by J. Scott Penberthy. George Ferguson added support for printing the AAAI copyright slug. Additional changes to aaai2026.sty and aaai2026.bst have been made by Francisco Cruz, Marc Pujol-Gonzalez, and Mico Loretan.

\bigskip
\noindent This work was supported by Beijing Natural Science Foundation L233033 and China Natural Science Foundation Youth Fund 62202267.

\bibliography{reference}

\end{document}